\def\year{2021}\relax
\documentclass[letterpaper]{article} 
\usepackage{aaai21}  
\usepackage{times}  
\usepackage{helvet} 
\usepackage{courier}  
\usepackage[hyphens]{url}  
\usepackage{graphicx} 
\usepackage{natbib}  
\usepackage{caption} 
\usepackage{booktabs}
\usepackage{graphicx}
\usepackage{amsmath}
\usepackage{amsthm}
\usepackage{booktabs}
\usepackage{algorithm}
\usepackage{algorithmic}
\usepackage{amsfonts,amssymb}
\usepackage{times}
\usepackage{epsfig}
\usepackage{graphicx}
\usepackage{amsmath}
\usepackage{amssymb}
\usepackage{multirow}
\usepackage{subfigure}
\usepackage{color}
\usepackage{booktabs}
\usepackage{subfigure}
\usepackage{graphicx}
\usepackage{times}
\usepackage{epsfig}
\usepackage{graphicx}
\usepackage{amsmath}
\usepackage{amssymb}
\usepackage{float}
\usepackage{caption}
\usepackage[switch]{lineno}
\title{Incremental Embedding Learning via Zero-Shot Translation}
\author{
	Kun Wei, Cheng Deng\footnote{Corresponding Author}, Xu Yang, and Maosen Li}

\affiliations{
	\\School of Electronic Engineering, Xidian University, Xi'an 710071, China\\
	\{weikunsk, chdeng.xd, xuyang.xd, maosenli95\}@gmail.com
}

\begin{document}

\maketitle

\begin{abstract}
	Modern deep learning methods have achieved great success in machine learning and computer vision fields by learning a set of pre-defined datasets.
	Howerver,  these methods perform unsatisfactorily when applied into real-world situations.
	The reason of this phenomenon is that learning new tasks leads  the trained model quickly forget the knowledge of old tasks, which is referred to as catastrophic forgetting.
	Current state-of-the-art incremental learning methods tackle catastrophic forgetting problem in traditional classification networks and ignore the problem  existing in embedding networks, which  are the basic networks for image retrieval, face recognition, zero-shot learning, etc.
	Different from traditional incremental classification networks, the semantic gap between the embedding spaces of two adjacent tasks is  the main challenge for embedding networks under incremental learning setting.
	Thus, we propose a novel class-incremental  method for embedding network, named as zero-shot translation class-incremental method (ZSTCI), which leverages zero-shot translation to estimate and compensate the semantic gap without any exemplars.
	Then, we try to learn a unified representation for two adjacent tasks in  sequential learning process, which captures  the relationships  of previous classes and current classes precisely.
	In addition, ZSTCI can easily be combined with existing regularization-based incremental learning methods to further improve performance of embedding networks.
	We conduct extensive experiments on CUB-200-2011 and CIFAR100, and the experiment results prove the effectiveness of our method.
	The code of our method has been released\footnote{https://github.com/Drkun/ZSTCI}.

\end{abstract}
\section{Introduction}
In recent years, incremental learning (IL)~\cite{mccloskey1989catastrophic, kirkpatrick2017overcoming} has gained significant attention in  machine learning~\cite{yang2018new} and computer vision fields~\cite{wang2018parameter}, which requests the model can learn new tasks sequentially  without forgetting the tasks learned previously.
Different from other traditional methods trained on a set of pre-defined datasets, incremental learning methods are trained in a consecutive manner.
For different tasks in the training process, only the data of current task is available to be learned.
Thus, in the process of sequential learning, the network would suffer from catastrophic forgetting~\cite{robins1995catastrophic, mccloskey1989catastrophic}, which means the network loses the knowledge learned from previous tasks.

\begin{figure}[!tb]
	\centering
	\subfigure[]{
		\centering
		\includegraphics[height=4.5cm]{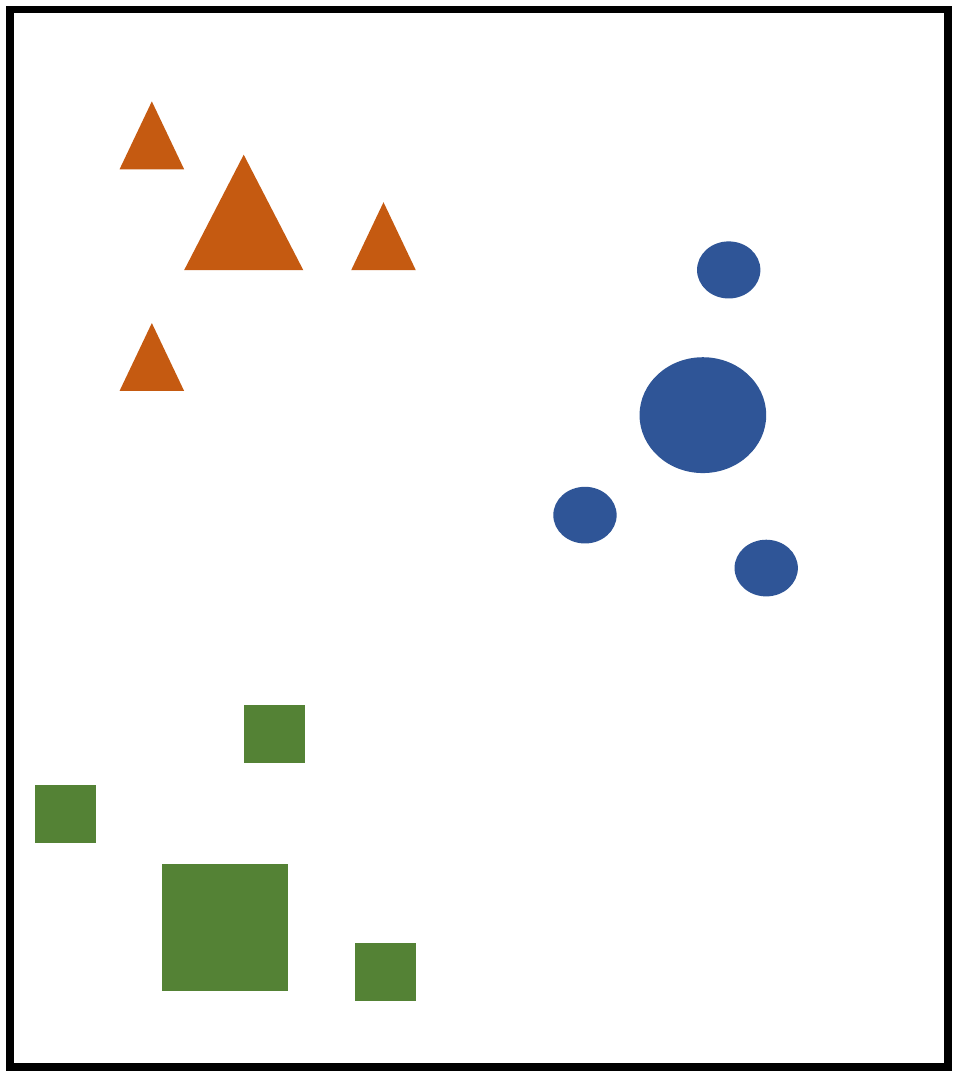}}
	\hspace{3pt}
	\subfigure[]{
		\centering
		\includegraphics[height=4.5cm]{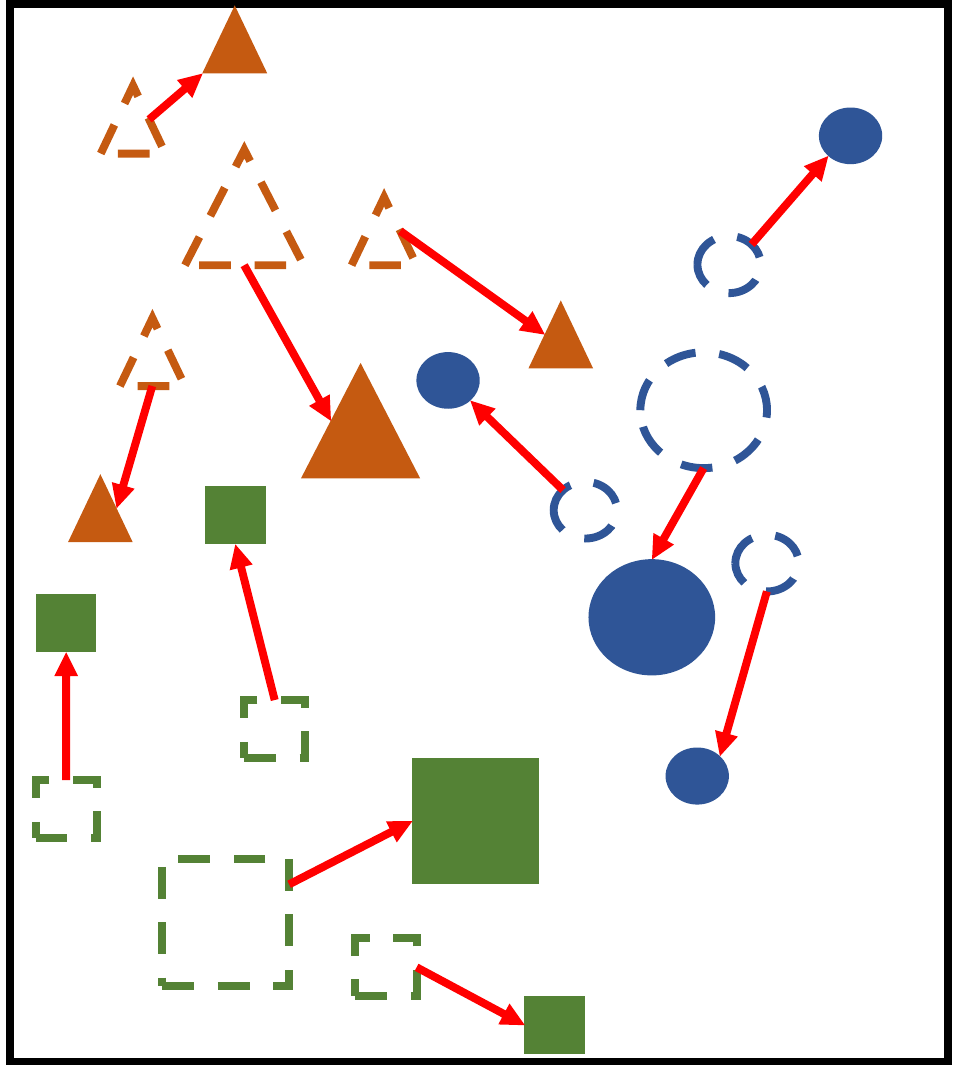}}
	
	\caption{Illustration of semantic gap. (a)  Data and prototypes of three classes of task 1 after training task 1.  (b) Data and prototypes of three classes after the training process of task 2.}
	\label{fig:intro}
\end{figure}

To alleviate catastrophic forgetting, many incremental learning strategies~\cite{liu2018rotate,rebuffi2017icarl} are proposed, which can be divided into three categories, storing training samples~\cite{rebuffi2017icarl,aljundi2018memory}, regularizing the parameters updates~\cite{li2017learning,liu2018rotate} and learning  generative models to replay the samples of previous tasks~\cite{shin2017continual,wu2018memory,seff2017continual}.
All these methods aims to transfer the knowledge of previous tasks to current task, which preserve the ability learned previously.
Learning without forgetting (LwF)~\cite{li2017learning} adds a distillation loss to preserve the old knowledge while sequentially learning new tasks.
iCaRL~\cite{rebuffi2017icarl} maintains an “episodic memory” of the exemplars and incrementally learns the nearest-neighbor classifier for new classes.
DGDMN~\cite{kamra2017deep}  uses Generative Adversarial Networks (GANs)~\cite{goodfellow2014generative} to generate old samples in each new phase for data replaying, and good results are obtained in the multi-task incremental setting.

The methods mentioned above aim to address catastrophic forgetting in classification networks, which learn the knowledge of new classes by adding new weights.
However, traditional incremental classification task tries to avoid the misclassification of old classes that caused by catastrophic forgetting. 
In contrast, we focus on the semantic gap in incremental embedding learning caused by catastrophic forgetting. 
The input data is directly mapped and regularized in the common embedding spaces by embedding networks without adding new weights, which are typically employed for hashing image retrieval~\cite{yang2020deep,yang2018shared,yang2017pairwise}, zero-shot recognition~\cite{chen2018zero}, domain adaptation~\cite{CSCL_Dong_ECCV2020,What_Transferred_Dong_CVPR2020}, clustering~\cite{yang2020adversarial,ZhangCSWD20,yang2019deep}, etc.
Different from traditional classification networks, the main reason leading to catastrophic forgetting is the semantic gap between two embedding spaces, which is led by the different classes of two adjacent tasks.
As shown in Figure~\ref{fig:intro}, data and prototypes, which are the mean latent features of the corresponding classes, is measured precisely after training task 1.
After the training process of task 2, the distribution between the data and prototypes become biased, leading to a unsatisfied performance.
SDC~\cite{yu2020semantic} notes and approximates the semantic gap of prototypes during training of new tasks, then  compensates the semantic gap without the need of any exemplars.
But SDC only estimates the semantic gap simply and ignores the relationship between the classes from different tasks, which influences the performance of semantic gap compensation. 

In this paper, we propose a novel method to alleviate catastrophic forgetting for embedding networks, named as zero-shot translation class-incremental method (ZSTCI).
We employ zero-shot translation to estimate the semantic gap between two adjacent embedding spaces, which leverages the generalization of network to achieve the translation of knowledge from seen classes to  unseen classes.
As for our task, the classes of previous tasks can be viewed as unseen classes for current task without any examples of previous tasks, and the classes of current task can be viewed as seen classes with the discriminative representation in two embedding spaces. 
Thus, zero-shot translation is employed to translate the prototypes of previous classes and current classes into a common embedding space to compensate the semantic gap between two embedding spaces.
In addition, we try to obtain a  unified representation for previous classes and current classes, which captures the discriminative relationships among these classes. 
Extensive experiments demonstrate that our method is able to alleviate catastrophic forgetting in embedding network effectively and can be flexibly combined with other incremental learning strategies to improve the performance further.

In summary, the contributions of this work are as follows:
\begin{itemize}
	\item  We construct a zero-shot translation model between two adjacent tasks, which  estimates  and compensates the semantic gap  between two adjacent embedding spaces.
	\item  We attempt to learn a unified representation in the common embedding space, which captures the relationships between previous classes and current classes precisely.
	\item  Extensive experimental results  demonstrate the effectiveness of our proposed method and our method can  be flexibly combined with existing regularization-based incremental learning methods, which can  further alleviate  catastrophic forgetting.
\end{itemize}
\begin{figure*}[tb!]
	\begin{center}
		\includegraphics[width=1\linewidth]{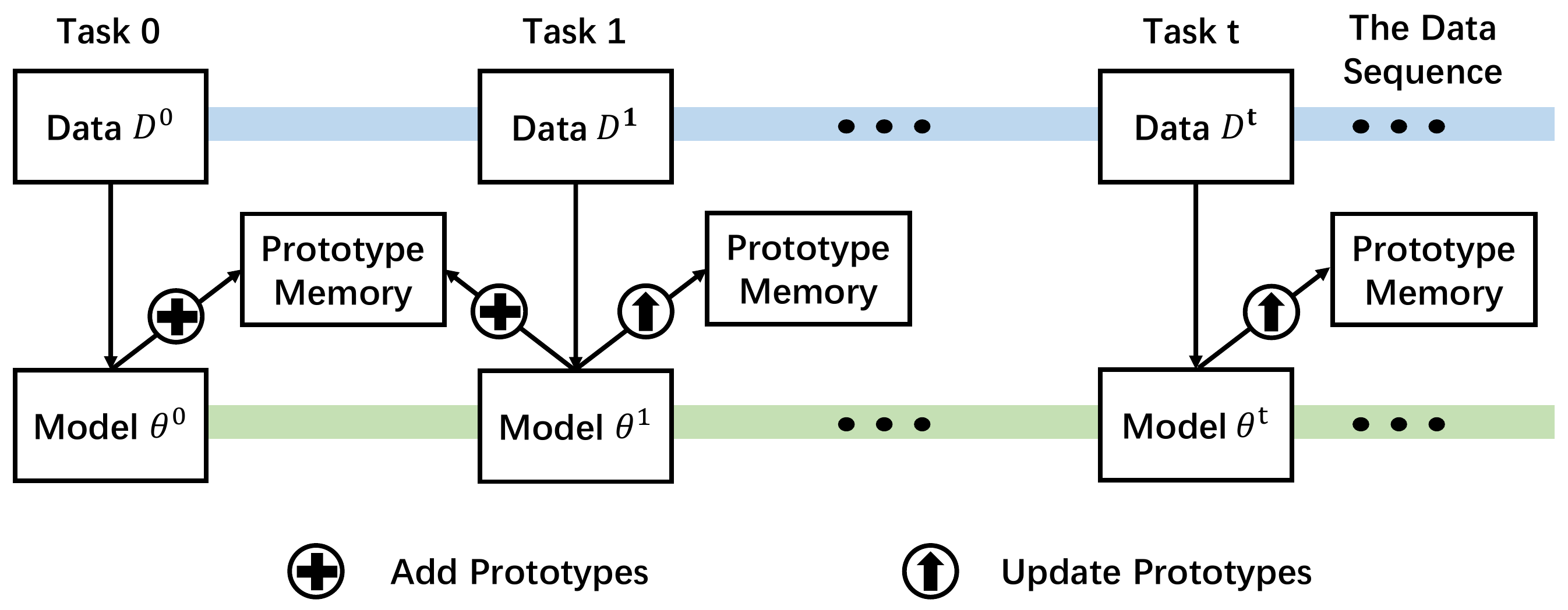}
	\end{center}
	
	\caption{The computing flow of the proposed zero-shot translation class-incremental method (ZSTCI). With learning different classes sequentially, the items in prototype memory are added and updated iteratively. }
	
	\label{fig:frame}
\end{figure*}
\section{Related work}
\paragraph{Incremental Learning.}
Incremental learning~\cite{mccloskey1989catastrophic, kirkpatrick2017overcoming}  is the learning pattern that requires the model has the ability to accumulate the knowledge of previous tasks  and capture the knowledge of current tasks simultaneously.
Catastrophic forgetting~\cite{robins1995catastrophic, mccloskey1989catastrophic} is the main reason to lead  the trained model  forgets the knowledge of previous tasks when new task arrives.
Many incremental learning methods have been proposed to alleviate the phenomenon of catastrophic forgetting.
These methods can be divided into three parts: regularizing the parameter updates~\cite{li2017learning,liu2018rotate}, which guarantees the outputs of previous network and current network are similar when given the same input, storing training samples of previous tasks~\cite{rebuffi2017icarl,aljundi2018memory}, which contains the discriminative knowledge of previous tasks, and training  generative networks to replay previous data~\cite{shin2017continual,wu2018memory,seff2017continual}, which converts incremental learning into traditional supervised learning.
Besides, SDC~\cite{yu2020semantic} is proposed to approximate the semantic drift after training new tasks, which is the complementary to several existing methods for incremental learning originally designed for classification networks.
Besides, incremental learning has been combined with other computer vision tasks, such as zero-shot learning~\cite{ijcai2020-77}, semantic segmentation~\cite{michieli2019incremental}, few-shot learning~\cite{tao2020few}, which brings the potential to bridge the semantic gap between the computer vision field and real-world situations.

Different from the methods mentioned above, our method leverages zero-shot translation to bridge the semantic gaps between different tasks and alleviate catastrophic forgetting.
In addition, our method tries to learning a unified representation for  previous tasks and  current task, which measures the relationships among classes precisely.
\paragraph{Zero-Shot Learning.}
Zero-shot learning (ZSL)~\cite{romera2015embarrassingly,chen2018zero,wei2019adversarial} is a hot topic in transfer learning, which handles issue that some test classes are not included in the training set.
The main solution is to leverage the generalization of network to transfer the knowledge from seen classes to unseen classes. 
Zero-shot learning method can be divided into two parts: embedding methods~\cite{socher2013zero,zhang2017learning}, which learn the connection between the visual and semantic space, generative  ZSL methods~\cite{xian2018feature,felix2018multi}, which leverage generative adversarial networks (GAN) models to generate discriminative features of unseen classes  and convert ZSL into traditional supervised learning.
A feature-generating network (f-CLSWGAN)~\cite{xian2018feature} is proposed by employing conditional Wasserstein GANs (WGANs)~\cite{arjovsky2017wasserstein}. 
Based on f-CLSWGAN, a new regularization is further employed~\cite{felix2018multi} for GAN training that forces the  generated visual features to reconstruct their original semantic embedding.
In addition, variational auto-encoder (VAE) is combined with GAN to synthesize more discriminative features, which obtains impressive performance~\cite{schonfeld2019generalized}.

Inspired by the methods mentioned above, we introduce zero-shot learning into class-incremental task, which bridges the semantic gap between two adjacent training tasks.

\section{Methodology}
To bridge the semantic gap between two tasks, we propose a novel class-incremental method for embedding networks, which employs zero-shot learning to estimate and compensate the semantic gap,  named as zero-shot translation class-incremental (ZSTCI).
Specifically, we  try to learn a  unified representation for the classes of  previous tasks and  current task, capturing the relationships between previous classes and current classes.
\subsection{Problem Formulation}
In this paper, we focus on the class-incremental embedding classification problem, where a network learns several tasks and the classes of theses tasks are not overlapped.
During the training process of task $t$, only one dataset $D^t$ is available, containing pairs $(x_{i},y_{i})$, where $x_i$ is an image of classes ${y_i} \in {C^t}$.
In addition, the number of pairs in $D^t$ is $n^t$.
${C^t} = \left\{ {c_1^t,c_2^t \ldots ,c_{{m^t}}^t} \right\}$ is a limited set of classes and $m^t$ is the number of classes in task $t$.
After the whole training process, we achieve the testing process on all classes $C = { \cup _i}{C^i}$.
Following the setting of other class-incremental methods, the task label at test time is not available.
\subsection{Incremental Learning for Embedding Network}
During the training process of task $t$, we employ embedding networks to  project images into a low-dimensional space, where the distances between different images  measured by $L2$-distance represent the similarity between the images.
The original images are mapped into the embedding space and regularized by triplet loss, which can be replaced as other objective functions in other embedding tasks.
The mapping process is noted as $z_i=F(x_i)$, where $x_i$ is the image data and $z_i$ is the latent feature in the embedding space.
The triplet loss ensures the anchor to be close to the positive sample and far from the negative one, which is formulated as:
\begin{equation}
{\mathcal{L}_{tri}} = \max \left( {0,{d_ + } - {d_ - } + m} \right),
\end{equation}
where $d_+$ ($d_-$)  is denoted as the $L2$-distance between the anchor and the positive sample (the negative sample).
In addition, $m$ is the margin value.
After the training process of embedding network, we denote the embedding space as the classification space and employ nearest class mean (NCM) classifier to achieve the classification, which can be denoted as:
\begin{equation}
c_j^ *  = \mathop {\arg \min }\limits_{c \in C} dist\left( {{z_j},{u_c}} \right),
\end{equation}
\begin{equation}
{u_c} = \frac{1}{{{n_c}}}\sum\limits_i {\left[ {{y_i} = c} \right]{z_i}},
\end{equation}
where $n_c$ is the number of training samples belonging to class $c$ and $[Q]=1$ if $Q$ is true, and 0 otherwise.
In addition, we denote $u_c$ as the prototype of class $c$.

When new tasks arrive, the embedding network is fine-tuned in the new datasets, and regularized by triplet loss.
Then, we employ the trained model to compute the prototypes of new classes and add  these prototypes into prototype memory, containing the prototypes for all learned classes.
Finally, we perform NCM for classification, which is denoted as embedding fine-tuning (E-FT).

\begin{figure}[tb!]
	\begin{center}
		\includegraphics[width=0.9\linewidth]{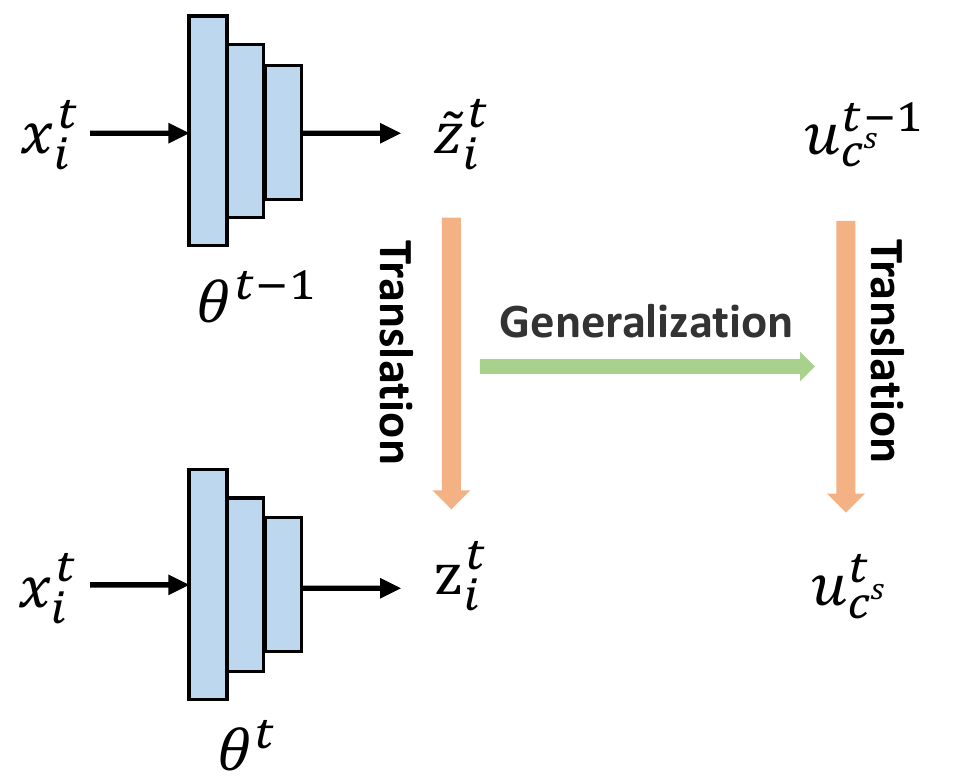}
	\end{center}
	
	\caption{The illustration of Zero-Shot Translation.}
	
	\label{fig:zero}
\end{figure}

For the embedding network, the objective function is denoted as:
\begin{equation}
\mathcal{L}_{emb}=\mathcal{L}_{tri-emb}.
\end{equation}

\subsection{Zero-Shot Translation}
The disjointness of classes in different tasks leads to the large semantic gap between the classification spaces.
Hence, the prototypes of previous classes stored in prototype memory are not compatible with new embedding network.
To bridge the gap and transfer the knowledge of previous classes, we need to transfer the prototypes of previous classes into the classification space of current task.
Different from traditional domain translation, the previous classes in source domain are not contained in the target classification space in class-incremental classification task, which limits the performance of  translation.
As shown in Figure~\ref{fig:zero},  we construct a zero-shot translation model to bridge two  different representations of the same input in two classification spaces, which leverages the generalization of the network to achieve the translation for the prototypes of previous classes.

\begin{figure}[tb!]
	\begin{center}
		\includegraphics[width=0.65\linewidth]{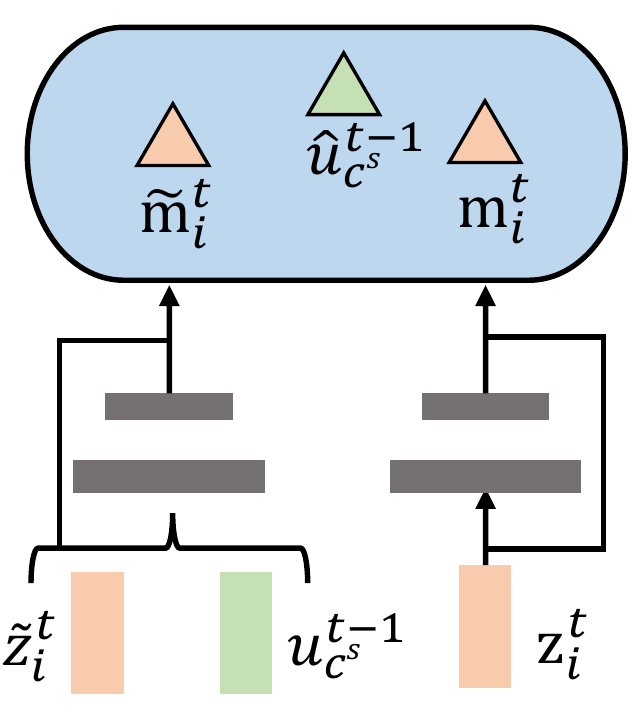}
	\end{center}
	\caption{The illustration of prototype update.}
	\label{fig:update}
\end{figure}

As shown in Figure~\ref{fig:frame}, after the training process of task $t$, we first add the prototypes $u_c^t$ of classes $C^t$ into prototype memory.
We refer to the  prototype means of previous class as $u_{c^s}^{t-1}$ $\left( {t > s} \right)$, which is the mean feature for class $c^s$ after the update in task $t-1$. 
Then we leverage zero-shot translation to construct a common embedding space to align the features from different classification spaces, as shown in Figure~\ref{fig:update}.
The latent features $\widetilde{z_i^t}$ and $z_i^t$ are extracted from the embedding network ${\theta ^{t-1}}$ and ${\theta ^t}$  given the same image input $x_i^t$.
Then, we project $\widetilde{z_i^t}$ and $z_i^t$ into a common embedding space by the zero-shot translation models $g_{old}$ and $g_{cur}$, denoted as $\widetilde{m}_i^t$ and $m_i^t$.
To preserve the representativeness and discrimination of the prototypes for each classes, we  leverage residual learning to achieve  zero-shot translation, which can be denoted as:
\begin{equation}
\widetilde{m_i^t} = \widetilde{z_i^t} + {g_{old}}\left( {\widetilde{z_i^t}} \right),
\end{equation}
\begin{equation}
m_i^t =z_i^t + {g_{cur}}\left( {z_i^t} \right).
\end{equation}
To align $\widetilde{m}_i^t$ and $m_i^t$, we design the align loss, which is denoted as:
\begin{equation}
\mathcal{L}_{align}=\frac{1}{{{n^t}}}\sum\limits_{i = 1}^{{n^t}} {{{\left\| {\widetilde{m}_i^t - m_i^t} \right\|}_1}}.
\label{eq:aglin}
\end{equation}
After the training process of the zero-shot translation models, we employ  $g_{old}$ to update $u_{c^s}^{t-1}$ and  $g_{cur}$ to update $u_{c^t}^{t}$.
Thus, the prototypes in prototype memory belong to the common embedding space.

\begin{table}[tb!]
	\large
	\centering
	\setlength{\tabcolsep}{2.4mm}{
		\scalebox{0.95}{
			\begin{tabular}{cccc}
				\toprule	
				Dataset & Classes & Image & Fine-grained\\		
				\midrule
				CUB & 200 & 11788 &  True\\
				CIFAR100 & 100 & 60000  & False \\
				\bottomrule
	\end{tabular}}}
	\caption{Datasets used in our experiments, and their statistics.}
	\label{tab:num_1}
\end{table}
\begin{table*}
	\LARGE
	\centering
	\setlength{\tabcolsep}{5mm}{
		\scalebox{0.72}{
			\begin{tabular}{cccccccccccc}
				\toprule
				Method& T1 &T2 & T3 & T4 & T5 & T6 & T7 & T8 & T9& T10\\
				\midrule
				E-FT & 88.9&75.3 & 68.4& 60.2& 57.4 &49.8 & 45.0& 39.6& 39.4&36.9\\
				E-FT+SDC  & 88.9 & 77.9 &72.2& 68.1 &65.9& 59.7&58.1& 54.7&52.1 &48.4\\
				E-FT+ZSTCI  & 88.9 &\textbf{81.5} & \textbf{75.8} &\textbf{71.6}& \textbf{67.4}&\textbf{62.5}& \textbf{61.0}& \textbf{58.6} &\textbf{55.9}&\textbf{52.1}\\
				\midrule
				E-LwF & 88.9& 78.0 & 72.5& 66.7& 61.9 &57.2 & 54.5& 53.4&49.5& 45.9\\
				E-LwF+SDC  & 88.9 & 79.0 & 72.8 &67.8& 61.4&57.1& 55.8& 54.4 &50.9& 48.4\\
				E-LwF+ZSTCI  & 88.9 & \textbf{80.1} & \textbf{73.7} &\textbf{69.9}& \textbf{64.4}&\textbf{61.2}& \textbf{59.4}& \textbf{58.2} &\textbf{56.9}& \textbf{55.0}\\
				\midrule
				E-EWC & 88.9& 76.9 & 68.3& 62.2& 60.0 &57.0 & 55.5& 51.0&53.3	& 51.1\\
				E-EWC+SDC  & 88.9 & 79.8 & 71.5 &67.0& 63.5& 60.7&59.5& 58.7& 56.5 &55.7\\
				E-EWC+ZSTCI  &88.9 & \textbf{80.1} & \textbf{72.7} &\textbf{68.9}& \textbf{65.8}&\textbf{61.9}&\textbf{61.2}& \textbf{59.9} &\textbf{59.0}& \textbf{58.1}\\
				\midrule
				E-MAS & 88.9& 74.7 & 66.4& 59.0& 58.2 &54.7 & 53.2& 48.7&50.1& 49.1\\
				E-MAS+SDC  & 88.9 & 76.2& 69.1 &64.1& 60.2&56.5& 55.7& 54.2 &53.0& 51.2\\
				E-MAS+ZSTCI  & 88.9 & \textbf{77.3} & \textbf{71.9} &\textbf{67.0}&\textbf{63.0}& \textbf{60.5}&\textbf{59.7}& \textbf{57.8}& \textbf{56.5} &\textbf{54.8}\\
				\bottomrule
	\end{tabular}}}
	
	\caption{The average incremental accuracy on CUB dataset.}	
	\label{tab: cub}
\end{table*}
\subsection{Unified Representation}
After zero-shot translation, the prototypes of current task and the previous tasks are represented in a common embedding space.
However, the distribution between these prototypes are not represented precisely, the reason of which is that these classes are  not learned and regularized simultaneously in the same space.
Thus, the goal of the translation network is to search a better embedding space to measure and regularize the samples of different classes precisely, compared with two adjacent embedding spaces. 
As shown in Figure~\ref{fig:update}, we design a unified representation strategy to learn these classes in a common embedding space, which  can not be completed in the classification space of current task.
The mapping model $g_{old}$  we employed projects not only the latent feature $\widetilde{z_i^t}$, but also the prototypes $u_{c^s}^{t-1}$ of previous classes  into the common embedding space. 
As for one latent feature $\widetilde{z_i^t}$, we select a prototype $u_{c^s}^{t-1}$ of  previous classes randomly in the training process.
Then, the triplet loss is leveraged to regularize the distribution of different classes, which is beneficial for obtaining a unified representation for all classes. 
In addition, we select the $m_i^t$, $\widetilde{m}_i^t$ and  $\widetilde{u}_{{c^s}}^s$ as the anchors respectively, which can be denoted as:
\begin{equation}
{\mathcal{L}_{tri-tran}} =\gamma \mathcal{L}_{tri}(m_i^t) + \beta \mathcal{L}_{tri}(\widetilde{m}_i^{t-1})+\delta \mathcal{L}_{tri}(\widetilde{u}_{{c^s}}^s),
\label{eq:tran}
\end{equation}
where $\gamma$, $\beta$, and $\delta$ are the hyper-parameters to weight the three triplet losses.

For the zero-shot translation model, the objective function can be denoted as:
\begin{equation}
{\mathcal{L}_{tran}} = \mathcal{L}_{tri-tran} +  \mathcal{L}_{align}.
\label{eq:all}
\end{equation}
\subsection{Training and Inference}
In the training stage, the model is trained sequentially on different tasks, the process of which is shown in Figure~\ref{fig:frame}.
The training processes of embedding network and zero-shot translation are iterative.
After the training process of embedding networks, which is regularized by $\mathcal{L}_{emb}$, we first add the prototypes of  new classes into prototype memory.
Then, we train zero-shot translation network to update the the prototypes of both new classes and old classes, which is regularized by $\mathcal{L}_{tran}$.
The parameters of $g_{old}$ and $g_{cur}$ are optimized jointly by Eq.~\ref{eq:all}, which is combined by Eq.~\ref{eq:aglin} and Eq.~\ref{eq:tran}. 
The $g_{old}$ and $g_{cur}$ constitute the translation network between two adjacent tasks. 

In the testing stage, we first map all the testing samples into the original classification space as latent features.
Then, all latent features of testing samples and class prototypes are projected into the common embedding space.
Finally, we will use NCM classifier for classification, which is defined as:
\begin{equation}
c_j^ *  = \mathop {\arg \min }\limits_{c \in C} dist\left( {{m_j},{u_c}} \right).
\end{equation}

\section{Experiment}
In this section, involved datasets, evaluation metrics and the implementation details are introduced.
Then, we will present the comparison results with several state-of-the-art incremental methods to prove the effectiveness of our proposed method.
Finally, the ablation studies will be presented to prove the effectiveness of different modules.

\paragraph{Datasets.}
We evaluate the methods on two popular datasets: CUB-200-2011 (CUB)~\cite{wah2011caltech} and CIFAR100~\cite{krizhevsky2009learning}.
Statistics of these datasets are presented in Table~\ref{tab:num_1}.
CUB is the typical dataset for many embedding tasks, such as zero-shot learning, fine-grained classification learning.
CIFAR100 is the typical dataset for class-incremental learning.
All these datasets are divided by classes into ten tasks randomly and the random seed is set as 1993.
\paragraph{Implementation Details.}
As for embedding network, ResNet-18~\cite{he2016deep} is selected as the backbone network pre-trained from ImageNet~\cite{deng2009imagenet} for CUB.
In addition, ResNet-32 is adopted for CIFAR100, which is without pre-training.
A triplet loss is employed to regularize the learning process of embedding network.
The training images are resized to $256\times 256$ for CUB and  $32\times 32$ for CIFAR100, then randomly cropped and flipped.
The epochs and learning rates are set to 50 and 1e-5 for CUB and CIFAR100 respectively.
The dimension of final embeddings normalized is 512.
All models are implemented  with Pytorch.
Adam optimizer~\cite{kingma2014adam} is employed to optimize the models and the batch size for all experiments is set to 32.

As for zero-shot translation network, the mapping models are two-layer fully-connected networks, and the dimension of hidden layer is 1024.
The epoch and batch size are set to 100 and 128 for CUB,  50 and 128 for CIFAR100.
In addition, the learning rate is set to 0.002 and the model is optimized by Adam optimizer.
$\gamma$, $\beta$ and $\delta$ are set to 1000, 100 and  100 for CUB, 200, 100 and 100 for CIFAR100 respectively. 

\begin{table*}
	\LARGE
	\centering
	\setlength{\tabcolsep}{5mm}{
		\scalebox{0.75}{
			\begin{tabular}{cccccccccccc}
				\toprule
				Method& T1 &T2 & T3 & T4 & T5 & T6 & T7 & T8 & T9& T10\\
				\midrule
				E-FT & 91.2& 72.4 & 65.0& 50.4& 46.1 &14.8 & 12.2& 10.2& 8.4&6.6\\
				E-FT+SDC  &91.2 & 74.4 &67.8& 55.3 &50.6& 17.0&13.9& 12.7&10.2 &8.1\\
				E-FT+ZSTCI  & 91.2 &\textbf{79.9} & \textbf{77.1} &\textbf{73.4}&\textbf{ 71.2}&\textbf{55.8}& \textbf{37.1}& \textbf{18.4} &\textbf{10.8}&\textbf{8.5}\\
				\midrule
				E-LwF & 91.2& 78.5& 76.7& 72.5& 70.6 &60.4 & 54.5& 49.6&44.5& 40.9\\
				E-LwF+SDC  & 91.2 & 78.6& 76.7 &72.7& 70.7&61.0& 55.5& 50.7&45.4& 42.0\\
				E-LwF+ZSTCI  & 91.2 & \textbf{78.8} & \textbf{77.1} &\textbf{73.1}& \textbf{71.1}&\textbf{62.3}&\textbf{ 57.4}&\textbf{53.2} &\textbf{49.0}& \textbf{46.1}\\
				\midrule
				E-EWC & 91.2& 78.5& 76.1& 73.0& 71.1 &59.6 & 50.6& 34.7&15.3& 10.3\\
				E-EWC+SDC  & 91.2& 78.5 & 76.1 &73.0& 71.3& 61.1&53.7& 43.4& 27.3 &14.4\\
				E-EWC+ZSTCI  &91.2 &\textbf{79.2} & \textbf{76.9}  &\textbf{73.6}&\textbf{71.8}& \textbf{62.6}&\textbf{57.6}&\textbf{53.4}& \textbf{48.4}&\textbf{43.3}& \\
				\midrule
				E-MAS & 91.2& 79.2 & 77.0 & 73.3& 70.7& 60.9 &53.7 & 40.8& 19.7&11.0\\
				E-MAS+SDC  & 91.2& \textbf{79.3}& 77.2 &73.3& 71.3&61.7& 55.2& 47.3 &32.2& 16.3\\
				E-MAS+ZSTCI  & 91.2 & \textbf{79.6} & \textbf{77.4} &\textbf{73.8}&\textbf{72.0}& \textbf{62.7}&\textbf{57.7}& \textbf{53.2}& \textbf{48.3} &\textbf{44.1}\\
				\bottomrule
	\end{tabular}}}
	\caption{ The average incremental accuracy on CIFAR100 dataset.}	
	\label{tab:cifar}
\end{table*}

\paragraph{Baseline Methods.}
\begin{itemize}
	\item \textbf{E-FT}: As described above.
	\item \textbf{E-LwF}~\cite{li2017learning}: It aims to guarantee the output embeddings $z_i^{t-1}$ of the models belonging to previous tasks is similar with the output embeddings $z_i^t$ of the current model when given the same input, which is achieved by constraining the parameters update. This leads to the following loss:
	\begin{equation}
	{\mathcal{L}_{LwF}} = \left\| {z_i^t - z_i^{t - 1}} \right\|,
	\end{equation}
	where $\left\| . \right\|$ refers to the  Frobenius norm.
	\item \textbf{E-EWC}~\cite{kirkpatrick2017overcoming}: It aims to retain the optimal parameters  of the former task during current training process. 
	The objective function of EWC is :
	\begin{equation}
	{\mathcal{L}_{EWC}} = \sum\limits_p {\frac{1}{2}} F_p^{t - 1}{\left( {\theta _p^t - \theta _p^{t - 1}} \right)^2},
	\end{equation}
	where ${F^{t - 1}}$ is the Fisher information matrix computed after the previous task $t-1$, and the summation goes over all parameters $\theta_p$ of the network.
	\item \textbf{E-MAS}~\cite{aljundi2018memory}: It aims to accumulate an importance measure for each parameter of the network based on how sensitive the predicted output function is to a change in this parameter.
	The objective loss is denoted as :
	\begin{equation}
	{\mathcal{L}_{MAS}} = \sum\limits_p {\frac{1}{2}{\Omega _p}{{\left( {\theta _p^t - \theta _p^{t - 1}} \right)}^2}},
	\end{equation}
	where ${\Omega _p}$ is  estimated by the sensitivity of the squared ${L_2}$ norm of the function output to their changes.
	
	These losses can be added to the metric learning loss to prevent forgetting while training embeddings continually:
	\begin{equation}
	\mathcal{L} = {\mathcal{L}_{ML}} + \gamma {L_C},
	\end{equation}
	where $C \in \left\{ {LwF,EWC,MAS} \right\}$, $\gamma$ is trade-off between the metric learning loss and the other losses, which is set to 1, 1e7 and 1e6 respectively.
	\item \textbf{SDC}~\cite{yu2020semantic}: It aims to  approximate the semantic drift of prototypes after training of new task.
	The method is complementary to several existing incremental learning methods to improve the performance further.
\end{itemize}

\paragraph{Evaluation Metric.}
We select  average incremental accuracy and average forgetting  as the evaluation metrics.
We denote ${a_{k,j}} \in \left[ {0,1} \right]$ as the accuracy of the $j$-th task ($j<k$) after training the network sequentially for $k$ tasks.
The average incremental accuracy at task $k$ is defined as ${A_k} = \frac{1}{k}\sum\limits_{j = 1}^k {{a_{k,j}}}$.

\begin{figure}[!tb]
	\centering
	\subfigure[CUB]{
		\centering
		\includegraphics[height=4.1 cm]{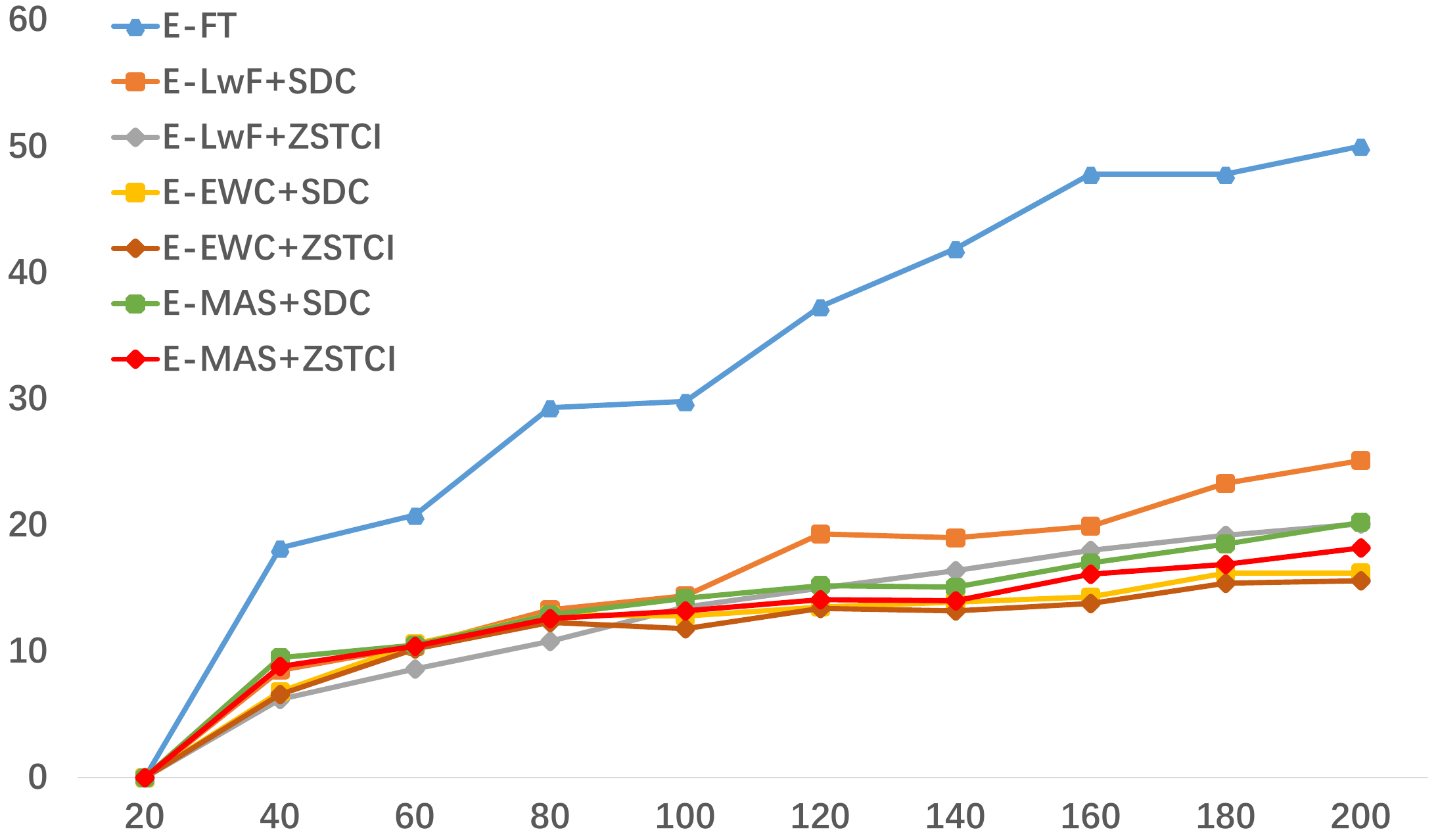}}
	\hspace{5pt}
	
	\subfigure[CIFAR100]{
		\centering
		\includegraphics[height=4.2cm]{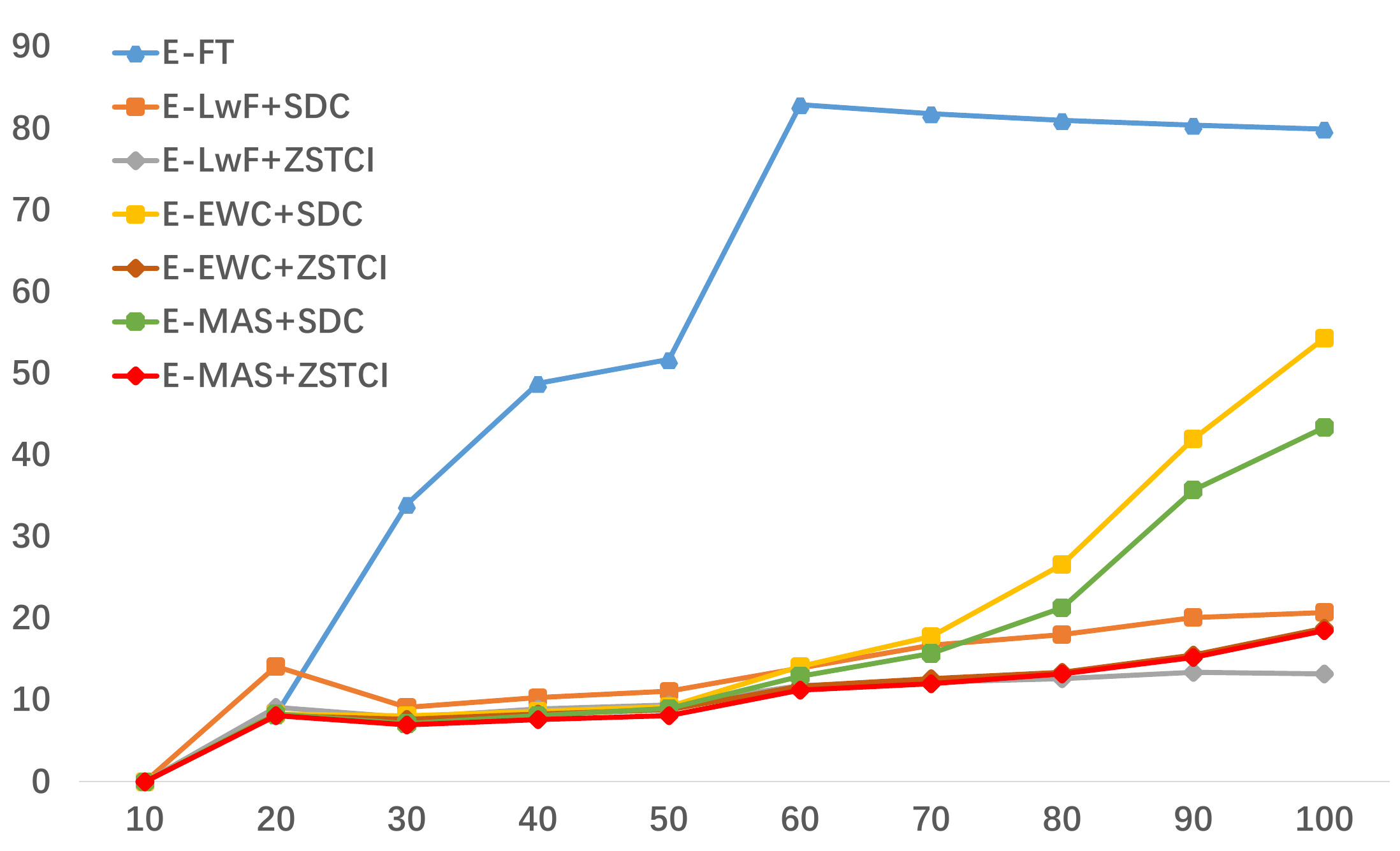}}
	
	\caption{Comparison of average forgetting with ten-task setting on CUB and CIFAR100 datasets.}
	\label{fig:forget}
\end{figure}

Average forgetting is defined to estimate the forgetting of previous tasks.
The forgetting for the $j$-th task is $f{}_j^k = \mathop {\max }\limits_{l \in 1, \ldots ,k - 1} \left( {{a_{l,j}} - {a_{k,j}}} \right),\forall j < k$.
The average forgetting at $k$-th task is written as ${F_k} = \frac{1}{{k - 1}}\sum\limits_{j = 1}^{k - 1} {f_j^k}$.
\subsection{Results and Analysis.} Table~\ref{tab: cub} and Table~\ref{tab:cifar} summarize the average incremental accuracy   results of all comparing methods and our method on CUB and CIFAR100 datasets.
We can note the methods equipped with ZSTCI obtain the best results in all tasks (except task1) on two datasets.
In addition, E-FT obtains the worst results on two datasets, which proves  the existence of catastrophic forgetting in embedding network.
The baseline methods (E-LwF/E-EWX/E-MAS) have the ability to alleviate the Catastrophic Forgetting in the learning process.
Based on these baseline methods, SDC and ZSTCT improve the ability of embedding network to alleviate Catastrophic Forgetting further.
On CUB, E-FT equipped with ZSTCI achieves 52.1\%, with 3.7\% improvements compared with E-FT equipped with SDC.
E-LwF equipped with ZSTCI achieves 55.0\%, with 6.6\% improvements compared with E-LwF equipped with SDC, 58.1\%, with 2.4\% improvements compared with E-EWC equipped with SDC, 54.8\%, with 3.6\% improvements compared with E-MAS equipped with SDC.
On CIFAR, E-FT equipped with ZSTCI achieves 8.5\%, with 0.4\% improvements compared with E-FT equipped with SDC.
E-LwF equipped with ZSTCI achieves 45.8\%, with 4.1\% improvements compared with E-LwF equipped with SDC, 34.0\%, with 28.9\% improvements compared with E-EWC equipped with SDC, 40.3\%, with 27.8\% improvements compared with E-MAS equipped with SDC.
The results prove ZSTCI is a better method to bridge  the semantic gap between two tasks compared with SDC and can easily be combined with  existing methods that prevent forgetting, such as EWC, LwF or MAS, to further improve the performance.

For CUB and CIFAR100, the average forgetting results are shown in Figure~\ref{fig:forget}.
With the increasing of classes,  the average forgetting becomes obvious for all methods, which proves  the existence of catastrophic forgetting in embedding network.
The methods combined with ZSTIC suffers from less forgetting than the methods combined with SDC on two datasets, which proves our method alleviates  catastrophic forgetting effectively.
For CIFAR100, the performance of alleviating catastrophic forgetting is more impressive compared with other baseline methods.

\begin{table}
	\LARGE
	\centering
	\setlength{\tabcolsep}{4mm}{
		\scalebox{0.72}{
			\begin{tabular}{ccccc}
				\toprule
				Method& E-FT &E-LwF & E-EWC & E-MAS \\
				\midrule
				Base  & 38.7& 44.3 & 51.8& 48.4\\
				+ZS  &\textbf{54.0} & 52.6 &56.9& 53.5\\
				+UR  &9.5 & 9.3 &9.3&  9.2\\
				+ZS+UR& 52.1&\textbf{55.0 }& \textbf{58.1} &\textbf{54.8}\\
				\bottomrule
	\end{tabular}}}
	\caption{ The average incremental accuracy on CUB.}	
	\label{tab:ab_cub}
\end{table}
\begin{table}
	\LARGE
	\centering
	\setlength{\tabcolsep}{4mm}{
		\scalebox{0.72}{
			\begin{tabular}{ccccc}
				\toprule
				Method& E-FT &E-LwF & E-EWC & E-MAS \\
				\midrule
				Base  & 6.6&40.9 & 10.3& 11.0\\
				+ZS  &7.4 & 45.8 &42.7& 43.2 \\
				+UR  &6.6 & 6.8 &6.5&  6.0\\
				+ZS+UR& \textbf{8.5} &\textbf{46.1} & \textbf{43.3} &\textbf{44.1}\\
				\bottomrule
	\end{tabular}}}
	\caption{ The average incremental accuracy on CIFAR100.}	
	\label{tab:ab_cifar}
\end{table}

\subsection{Ablation Study}
We conduct one groups of ablation experiments to study the effectiveness of our method.

The results of our basic model added different modules are present in Table~\ref{tab:ab_cub} and Table~\ref{tab:ab_cifar}.
The basic model is embedding model equipped with some Incremental Learning methods, such as E-LwF, E-EWC and E-MAS.
Based on the base model, we add Zero-Shot translation modules and  unified representation strategy, which are represented as “ZS” and “UR” respectively. 
The improvement of adding ``ZS'' indicates the Zero-Shot translation model estimate and compensate the semantic gap effectively.
When only adding ``UR'' into the base model, all the samples of precious classes cannot be projected into the common embedding space and be misclassified and only the classes of current task are classified precisely. 
When adding ``ZS'' and ``UR'' into the base model, the performance of model improve further,  which proves unified representation strategy can capture the distribution between the classes precisely.

\section{Conclusions}
In this paper, we propose a novel class-incremental method to alleviate catastrophic forgetting for embedding network.
To estimate and compensate the semantic gap between the classification space of two adjacent tasks, we construct a zero-shot translation model to map the prototypes into a common embedding space, where the latent features from two domains are aligned and measured precisely.
Then, we aims to obtain a unified representation for all classes, which can capture and measure the distribution between the classes.
In addition, our proposed method can flexibly be combined with other regularization-based incremental learning methods to improve the performance further.
Experiments show that our method outperforms previous methods by a large margin on two benchmark datasets.

\section*{Acknowledgments}
Our work was supported in part by the National Natural Science Foundation of China under Grant 62071361, and the National Key R\&D Program of China under Grant 2017YFE0104100.

\bibliography{ref}
\bibliographystyle{aaai21}

\end{document}